\documentclass[conference]{IEEEtran}
\IEEEoverridecommandlockouts
\usepackage{cite}

\usepackage{amsmath,amssymb,amsfonts}
\usepackage{algorithmic}
\usepackage{graphicx}   
\usepackage{textcomp}
\usepackage{xcolor}
\usepackage[hidelinks]{hyperref}
\usepackage{stfloats}
\usepackage{booktabs, tabularx}
\usepackage{rotating}
\usepackage{longtable}
\usepackage{soul}
\usepackage{subfigure}
\usepackage{url}
\usepackage{float}
\usepackage{array}
\usepackage{multirow}
\usepackage{ragged2e}
\usepackage{tabularray}
\usepackage{orcidlink}  

\usepackage [english]{babel}
\usepackage [autostyle, english = american]{csquotes}
\MakeOuterQuote{"}

\def\BibTeX{{\rm B\kern-.05em{\sc i\kern-.025em b}\kern-.08em
    T\kern-.1667em\lower.7ex\hbox{E}\kern-.125emX}}

\newcommand{\refappendix}[1]{\hyperref[#1]{Appendix~\ref*{#1}}}

\makeatletter
\newcommand{\linebreakand}{%
  \end{@IEEEauthorhalign}
  \hfill\mbox{}\par
  \mbox{}\hfill\begin{@IEEEauthorhalign}
}
\makeatother

\usepackage{soul}

\begin{document}

\bstctlcite{IEEEexample:BSTcontrol} 



\title{Robustness Verification of an Autonomous Underwater Vehicle-based Plankton Classifier\\ 
\thanks{This project has received funding from the European Union’s Horizon 2020 research and innovation programme under the Marie Sklodowska-Curie COFUND grant agreement no. 101034248. This work is partly funded by SFI HARVEST, 309661 by the Research Council of Norway.}
}

\author{
  \IEEEauthorblockN{\parbox{0.45\textwidth}{\centering 
    Abdelrahman Sayed Sayed\orcidlink{0000-0002-8912-0679}
  }}
  \IEEEauthorblockA{\parbox{0.45\textwidth}{\centering 
    \textit{Univ Gustave Eiffel, COSYS-ESTAS} \\
    F-59657 Villeneuve d’Ascq, France\\
    abdelrahman.ibrahim@univ-eiffel.fr
  }}
  
  \and
  
  \IEEEauthorblockN{\parbox{0.45\textwidth}{\centering 
    Pierre-Jean Meyer\orcidlink{0000-0002-8167-3156}
  }}
  \IEEEauthorblockA{\parbox{0.45\textwidth}{\centering 
    \textit{Univ Gustave Eiffel, COSYS-ESTAS} \\
    F-59657 Villeneuve d’Ascq, France\\
    pierre-jean.meyer@univ-eiffel.fr
  }}

  \linebreakand 

  \IEEEauthorblockN{\parbox{0.45\textwidth}{\centering 
    Asgeir J. Sørensen\orcidlink{0000-0002-7078-0298}
  }}
  \IEEEauthorblockA{\parbox{0.45\textwidth}{\centering 
    \textit{Department of Marine Technology} \\
    \textit{Norwegian University of Science and Technology (NTNU)}\\
    Trondheim, Norway \\
    asgeir.sorensen@ntnu.no
  }}
  
  \and
  
  \IEEEauthorblockN{\parbox{0.45\textwidth}{\centering 
    Mohamed Ghazel\orcidlink{0000-0002-1160-7997}
  }}
  \IEEEauthorblockA{\parbox{0.45\textwidth}{\centering 
    \textit{Univ Gustave Eiffel, COSYS-ESTAS} \\
    F-59657 Villeneuve d’Ascq, France\\
    mohamed.ghazel@univ-eiffel.fr
  }}
}

\maketitle

\begin{abstract}

The assessment of planktonic standing stocks and microorganism structures is critical for understanding upper ocean biological processes. Currently, autonomous underwater vehicles (AUVs) equipped with in-situ optical imaging and artificial intelligence (AI) methods offer a promising solution for persistent surveillance, mapping and monitoring of planktonic life. However, current AI methods often lack robustness in dynamic, unstructured environments, where environmental noise and non-biological artifacts lead to frequent misclassifications. Standard convolutional neural network (CNN) classifiers often struggle with such conditions, leading to misclassifications that require time-consuming manual validation by marine biologists.

To address this issue, we propose a novel robustness verification framework for in-situ plankton classifiers based on reachability analysis. We also introduce a continuous-time neural ordinary differential equation (neural ODE) classification model leveraging the high-resolution imaging capabilities of the SilCam particle imager. In this paper, we demonstrate the effectiveness of the proposed framework by formally verifying the robustness of the neural ODE model against environmental perturbations. We demonstrate that our verification framework acts as an automated filter providing formal guarantees of model stability against ambiguous data, thereby improving the reliability of autonomous sampling and reducing the post-processing workload.
\end{abstract}

\begin{IEEEkeywords}
Neural ODE, CNN, Reachability analysis, Formal methods, Robustness verification, AUVs
\end{IEEEkeywords}

\section{Introduction}
\label{sect:intro}

The growing interest in assessing and monitoring planktonic standing stocks and microorganism structures in the upper water column (mesopelagic zone) is critical for understanding the impact of climate change on ocean processes. To achieve continuous systematic ecosystem surveillance and monitoring, the oceanographic community is increasingly relying on autonomous underwater vehicles (AUVs) equipped with advanced optical sensors, such as the SilCam \cite{davies2017use}, and Artificial Intelligence (AI)-based tools. These AUVs enable high-resolution imaging and intelligent onboard sampling \cite{saad2020advancing}, significantly accelerating the detection and classification of microorganisms while simultaneously reducing the operational cost of deploying fleets of research vessels. Currently, the automated classification of in-situ plankton imagery is mostly dependent on convolutional neural networks (CNNs), including task-specific architectures such as ZooplanktoNet \cite{dai2016zooplanktonet}, and in-situ imaging and analysis pipelines built around the SilCam imager \cite{davies2017use} and PyOPIA toolbox \cite{davies2023pyopia}.
However, deploying AI-based classification models in uncertain and unstructured underwater environments raises significant challenges. Compared to controlled laboratory settings, in-situ ocean imaging is affected by dynamic environmental noise, turbidity, and particularly by gas bubbles generated by the AUV's own motion. These appear as bright artifacts overlaid on the imaged particles and, hence, routinely trigger misclassifications.

To bridge the gap between reliable autonomy and AI-based classification models whose behaviour can be trusted, there is a need for rigorous verification tools that can certify the stability of classifiers against such environmental perturbations. While current CNN-based classifiers are static models often vulnerable to adversarial perturbations \cite{szegedy2014intriguing,goodfellow2015explaining}, neural ordinary differential equations (neural ODE) \cite{chen2018neural} have emerged as a powerful and robust alternative. Modeled as a continuous-time system, they have been empirically shown to be naturally more robust against some adversarial attacks \cite{carrara2019robustness,carrara2021defending,yan2022robustnessneuralordinarydifferential,luo2024stable}. 

Beyond such empirical robustness, formal verification provides provable guarantees on the classifier's behaviour over an entire set of perturbed inputs. Among well-known formal methods, reachability-based techniques propagate an input set layer-by-layer to over-approximate the reachable outputs \cite{meyer2021interval}. However, recent tools target mostly feed-forward ReLU networks, and certifying the continuous-time dynamics of a neural ODE remains mostly unexplored \cite{sayed2025bridgingneuraloderesnet}.

In this work, we propose a novel robustness verification framework for AUV-based plankton classifiers. Namely, we leverage the robustness properties of neural ODE and reachability analysis to formally guarantee the stability of classification results against specific noise thresholds. By integrating this verifier into the AUV processing pipeline, we aim to automatically filter ambiguous data and misclassifications, thereby enhancing the reliability of in-situ ocean observation. Section~\ref{Sect:node} first introduces the neural ODE model class we consider. Section~\ref{Sect:verifier_framework} presents the verification framework, Section~\ref{Sect:experimental_setup} describes the experimental setup and problem definition, and Section~\ref{Sect:results} reports the verification results.

\section{Neural Ordinary Differential Equations}
\label{Sect:node}

A neural ODE \cite{chen2018neural} models a continuous-depth transformation of a hidden state $z(t) \in \mathbb{R}^{n}$ as the solution to an initial value problem

\begin{equation}
    \frac{dz(t)}{dt} = f\big(z(t), t, \theta\big), \qquad z(t_0) = z_0,
    \label{eq:node}
\end{equation}
where the vector field $f$ is a neural network that is parameterized by trainable weights $\theta$. The model output is obtained by integrating \eqref{eq:node} from $t_0$ to a final time $t_f$,
\begin{equation*}
    z(t_f) = z_0 + \int_{t_0}^{t_f} f\big(z(t), t, \theta\big)\, dt,
    \label{eq:node_integral}
\end{equation*}
which is evaluated numerically with an ODE solver. Replacing the discrete stack of residual layers of a standard neural network by continuous dynamics \eqref{eq:node} is what gives neural ODE their smooth and stability-related robustness properties \cite{yan2022robustnessneuralordinarydifferential}.
In practice, the ODE block does not act on the raw input but is embedded between neural network layers, yielding what we refer to as a \emph{general neural ODE} (GNODE). In a formal way, a GNODE classifier is the composition
\begin{equation}
    \mathcal{N} = g_{\mathrm{post}} \circ \Phi \circ g_{\mathrm{pre}} : \mathbb{R}^{m} \to \mathbb{R}^{c},
    \label{eq:gnode_arch}
\end{equation}
where $g_{\mathrm{pre}} : \mathbb{R}^{m} \to \mathbb{R}^{n}$ is a feature-extraction network mapping an input (an image in our case) to the ODE initial state $z_0 = g_{\mathrm{pre}}(x)$, $\Phi : z_0 \mapsto z(t_f)$ is the flow obtained by integrating \eqref{eq:node}, and $g_{\mathrm{post}} : \mathbb{R}^{n} \to \mathbb{R}^{c}$ is a classification head mapping the terminal state to the $c$ class scores.

We consider the autonomous GNODE, where the vector field $f$ is a feed-forward network independent of $t$. In our model, $f$ is a single hidden-layer network
\begin{equation}
    \frac{dz(t)}{dt} = W_2\, \sigma\!\big(W_1 z(t) + b_1\big) + b_2,
    \label{eq:gnode}
\end{equation}
where $\sigma$ is an element-wise activation function and $W_1, W_2, b_1, b_2$ are the trainable parameters of the ODE block. In the architecture used throughout this paper (Section~\ref{Sect:experimental_setup}), the state dimension is $n = 32$ and $\sigma$ is either the hyperbolic tangent (Tanh) or the rectified linear unit (ReLU) activation function.

\begin{figure*}[t]
\centerline{\includegraphics[width=0.45\textwidth]{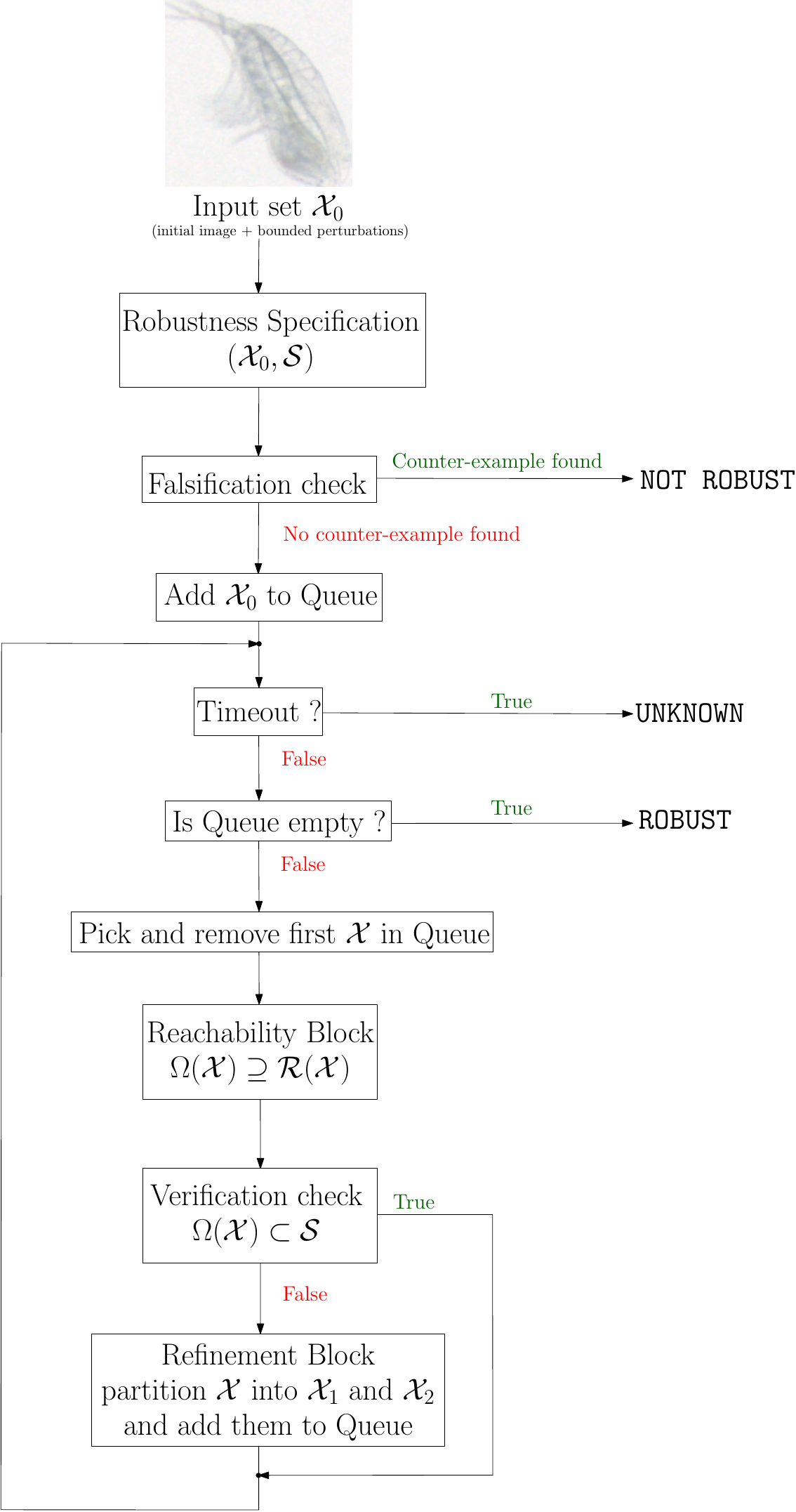}}
\caption{Verifier Architecture}
\label{fig:verif_arch}
\end{figure*}

\section{Underwater Image Verification Framework} 
\label{Sect:verifier_framework}

The architecture of the proposed verifier is illustrated in Figure~\ref{fig:verif_arch}. The framework is designed to formally guarantee that a classified image from in-situ images remains robust against environmental or user induced perturbations. This process is divided into two main phases: an initial falsification check and a rigorous verification and refinement loop.

\subsection{Initialization and Robustness Specification}

First, we define an initial input set $\mathcal{X}_{0}$ (representing an in-situ image with perturbation noise) and a desired classification $\mathcal{S}$ (the region of the output space corresponding to the correct class label). The goal is to prove that for all $x \in \mathcal{X}_0$, the classified output of the model remains within $\mathcal{S}$.
 
\subsection{Image Falsification Check}

This preliminary check avoids the unnecessary computational cost of full reachability analysis. The verifier generates $1000$ random samples from the entire input set $\mathcal{X}_0$ (i.e., the set of all potential noisy variations of the image) and evaluates the neural ODE classifier against the specification. This sample count was set empirically across the examples presented in the TNODEV toolbox \cite{sayed2026tnodevtoolboxneuralode}, and is large enough to reliably expose specification-violating inputs in this pre-check. The outcome of this check determines the next step:
\begin{itemize}
    \item Counter-example found: if any sample violates the desired classification $\mathcal{S}$, the verification process terminates,  returning a \texttt{NOT ROBUST} verdict, indicating that the classifier is not robust for this image.
    \item No counter-example: if no violation is detected among the samples, the verifier proceeds to the verification and refinement loop.
\end{itemize}

\subsection{Image Verification and Refinement}

In this phase, the verifier iteratively proves that the robustness specification $\mathcal{S}$ is satisfied for the input image set $\mathcal{X}_0$ using reachability analysis. As illustrated in Figure~\ref{fig:verif_arch}, this loop consists of three main blocks:
\begin{enumerate}
    \item Reachability Block: the verifier propagates the input set $\mathcal{X}_0$ through the layers of the neural ODE classifier to compute a guaranteed over-approximation $\Omega(\mathcal{X}_0) \supseteq \mathcal{R}(\mathcal{X}_0)$ of the exact output reachable set $\mathcal{R}(\mathcal{X}_0) = \{\mathcal{N}(x) : x \in \mathcal{X}_0\}$. The ODE block is handled with continuous time mixed-monotonicity reachability, as discussed in \cite{sayed2025mixed}, while the surrounding pre-ODE convolutional and post-ODE fully-connected layers are propagated with the star-set method of NNV 2.0 \cite{lopez2023nnv} in a hybrid pipeline.
    \item Verification Check: the resulting over-approximation is checked against $\mathcal{S}$, and if $\Omega(\mathcal{X}_0) \subset \mathcal{S}$ (i.e., the correct class score remains dominant over the entire set), the current image subset is verified.
    \item Refinement Block: if the check returns false, the input set is partitioned into smaller subsets (in our case, shaped as intervals, i.e.\ axis-aligned hyperrectangles), which will each go again through the reachability and verification steps 1 and 2 described above. This partitioning uses heuristics that identify and split the input dimensions that contribute most to the output uncertainty \cite{sayed2026tnodevtoolboxneuralode}.
\end{enumerate}

The verifier repeats this loop, refining the unverified subsets, until either every subset satisfies the specification yielding a \texttt{ROBUST} verdict, or the algorithm times out and thus returns the verdict \texttt{UNKNOWN}.
An \texttt{UNKNOWN} verdict is not a failure of the method, but simply the standard inconclusive outcome of sound but incomplete reachability-based verifiers such as NNV 2.0 \cite{lopez2023nnv}, as the over-approximation is too loose to produce a verdict on the robustness specification within the allocated computation time limit. We discuss its operational interpretation in Section \ref{Sect:results}.

\section{Experimental Setup} 
\label{Sect:experimental_setup}

The neural ODE networks were trained in PyTorch using the \texttt{torchdiffeq} adjoint ODE solver \cite{torchdiffeq}, on a desktop with an AMD Ryzen\texttrademark~9 5950X CPU (16 cores, 32 threads), 96~GB of RAM, and a NVIDIA GeForce RTX~5080 GPU with 16~GB of VRAM. The trained neural ODE models are available at the following repository: \url{https://github.com/ab-sayed/Robustness-Verification-of-an-AUV-based-Plankton-Classifier}

\subsection{Underwater Plankton Dataset}
 
We used the in-situ SilCam images collected through the PyOPIA ocean particle image analysis toolbox \cite{davies2023pyopia}. The dataset contains $7$ plankton classes (bubble, copepod, diatom chains, fecal pellets, oil, oily gas particles, and other) for a total of $7738$ images, distributed as $2636$ bubbles, $657$ copepods, $850$ diatom chains, $514$ fecal pellets, $671$ oil, $479$ oily gas particles, and $1931$ other. Although copepods represent only a small fraction of the dataset ($657$ images), the neural ODE classifier still achieves strong copepod classification recognition of $81.1\%$. We attribute this in part to the parameter efficiency of neural ODE as their weight-sharing, continuous-depth formulation uses far fewer parameters compared to an equivalent discrete network \cite{chen2018neural}, which reduces the risk of overfitting on under-represented classes.

\subsection{Classifier Architectures}

\begin{figure*}[t]
\centerline{\includegraphics[width=\textwidth]{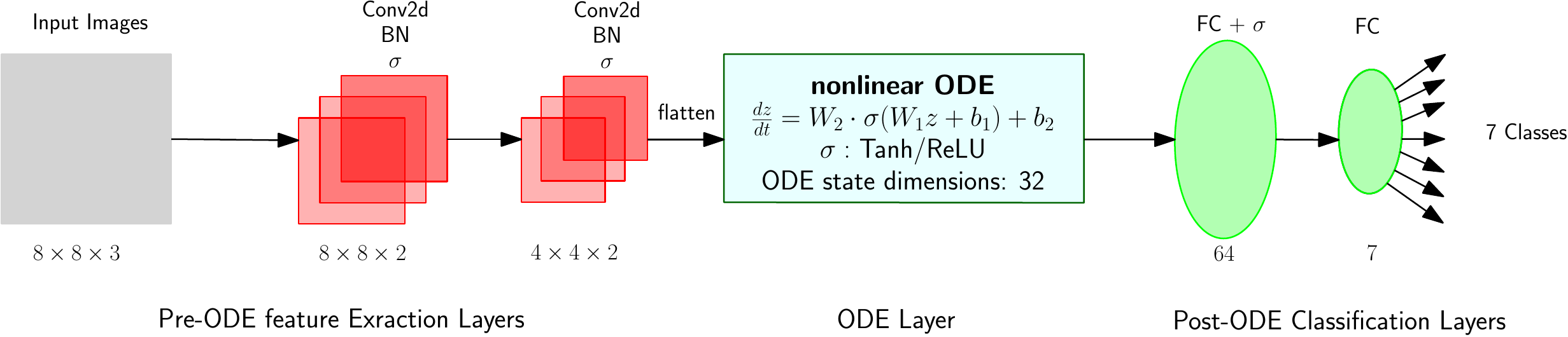}}
\caption{Neural ODE model architecture}
\label{fig:verif_SFI_Harvest}
\end{figure*}

Our neural ODE classifier illustrated in Figure \ref{fig:verif_SFI_Harvest} follows the in-situ plankton classification architecture of the CNN classifier introduced in \cite{saad2020advancing}, but adopts a compact and verifiable architecture in which the deep convolutional stack is reduced and the core nonlinear transformation is realized by a neural ODE block. From an input image $8\times8\times3$ ($192$ dimension), two pre-ODE feature-extraction blocks (2D convolution, batch normalization, and activation function $\sigma$) reduce the input to a flattened $32$-dimensional feature vector. This vector is the initial state $z_0$ of the ODE layer defined in \eqref{eq:gnode} with state dimension $n = 32$. The integrated state $z(t_f)$ is then passed to a post-ODE classification block of two fully-connected layers ($64$ then $7$ units), producing the score over the $7$ classes. The reduced input resolution and compact ODE state are deliberate choices made
primarily for verification tractability, as a smaller input dimension ($8\times8\times3 = 192$) and state dimension ($n = 32$) yield tighter reachable set over-approximations and fewer dimensions to split during refinement, which together lower the verification cost. These dimensions are not the product of a heuristic architecture search, as they follow from a direct compaction of the original CNN classifier \cite{saad2020advancing}, reducing its convolutional stack and replacing the core nonlinear transformation with the neural ODE block.

We evaluate two variants of this architecture that share the same structure and differ only in the activation function $\sigma$ used in the feature-extraction, ODE, and post-ODE layers:

\subsubsection{Neural ODE-based classifier with ReLU}
The variant using the ReLU activation, $\sigma(s) = \max(0, s)$.

\subsubsection{Neural ODE-based classifier with Tanh}
The variant using the Tanh activation, $\sigma(s) = \tanh(s)$.

\subsection{Problem Definition}
\label{subsec:problem_defin}

Let $\mathcal{N} : \mathbb{R}^{m} \to \mathbb{R}^{c}$ denote the trained neural ODE classifier mapping an input image to class scores, with input values $m = 8\times 8\times 3 = 192$ and classes $c = 7$. Let $x^\star \in \mathbb{R}^{m}$ be a nominal image that is \emph{correctly} classified, with true label
\begin{equation*}
    \ell = \arg\max_{j \in \{1,\dots,c\}} \mathcal{N}_j(x^\star).
\end{equation*}

\paragraph{Perturbation strategy}
We model environmental noise as a bounded $L_\infty$ perturbation applied to a subset $P \subseteq \{1,\dots,m\}$ of $k = |P|$ pixels, with magnitude $\varepsilon$ expressed in normalized intensity units (i.e., $\varepsilon = \alpha/255$). This induces the input set
\begin{equation*}
    \mathcal{X}_0 = \Big\{ x \in \mathbb{R}^{m} \;:\; \forall i \in P, |x_i - x^\star_i| \le \varepsilon; \forall i \notin P , x_i = x^\star_i \ \Big\},
    \label{eq:input_set}
\end{equation*}
which is an axis-aligned box (hyperrectangle). We consider $k \in \{1, \ldots, 192\}$ attacked pixels, ranging from a single-pixel perturbation to the full $192$-pixel image, over a range of $L_\infty$ radii $\varepsilon$.

\paragraph{Robustness specification}
Robustness requires that the correct label $\ell$ remain dominant over the entire perturbation set. The safe output region is
\begin{equation*}
    \mathcal{S} = \Big\{ y \in \mathbb{R}^{c} \;:\; y_\ell > y_j \ \ \forall j \neq \ell \Big\}.
\end{equation*}

\paragraph{Verification goal}
The classifier is certified robust for $x^\star$ under perturbation $(\varepsilon, P)$ if and only if every reachable output keeps $\ell$ dominant, i.e.
\begin{equation*}
    \mathcal{N}(\mathcal{X}_0) \subseteq \mathcal{S}.
\end{equation*}

The verifier in Section~\ref{Sect:verifier_framework} returns \texttt{ROBUST} when this inclusion is proven, \texttt{NOT ROBUST} when a counter-example is found by the falsification check, and \texttt{UNKNOWN} when neither verdict is established within the allocated time limit.

\section{Verification Results}
\label{Sect:results}

Tables \ref{tab:verification_results_RELU_AF} and \ref{tab:verification_results_Tanh_AF} report the verification verdicts and runtimes for the ReLU and Tanh variants across different perturbation settings as mentioned in Section~\ref{subsec:problem_defin}. 

\begin{table}[htbp]
    \centering
    \caption{Robustness verification results for neural ODE-based classifier with ReLU activation}
    \label{tab:verification_results_RELU_AF}
    \renewcommand{\arraystretch}{1.3}
    \resizebox{\columnwidth}{!}{%
      \begin{tabular}{@{}cccc@{}} 
        \toprule
        \textbf{Pixels Attacked} & \textbf{Noise ($L_{\infty}$)} & \textbf{Verdict} & \textbf{Time} \\
        \midrule
        1 / 192   & 0.1/255  & \texttt{ROBUST}                & 128.1 s \\
        1 / 192   & 0.5/255  & \texttt{ROBUST}                & 104.2 s \\
        10 / 192  & 0.1/255  & \texttt{ROBUST}                & 104.8 s \\
        192 / 192 & 0.01/255 & \texttt{ROBUST}                & 108.9 s \\
        192 / 192 & 5/255    & \texttt{NOT ROBUST} & 0 s     \\
        192 / 192 & 10/255   & \texttt{NOT ROBUST} & 0 s     \\
        192 / 192 & 1/255    & \texttt{Unknown}               & \bfseries\colorbox{pink}{7200 s}  \\
        192 / 192 & 0.05/255 & \texttt{Unknown}              & \bfseries\colorbox{pink}{7200 s}  \\
        \bottomrule
      \end{tabular}%
    }
\end{table}

\begin{table}[htbp]
    \centering
    \caption{Robustness verification results for neural ODE-based classifier with Tanh activation}
    \label{tab:verification_results_Tanh_AF}
    \renewcommand{\arraystretch}{1.3}
    \resizebox{\columnwidth}{!}{%
      \begin{tabular}{@{}cccc@{}} 
        \toprule
        \textbf{Pixels Attacked} & \textbf{Noise ($L_{\infty}$)} & \textbf{Verdict} & \textbf{Time} \\
        \midrule
        1 / 192   & 0.1/255  & \texttt{ROBUST}                & 119.2 s \\
        1 / 192   & 0.5/255  & \texttt{ROBUST}                & 101.4 s \\
        10 / 192  & 0.1/255  & \texttt{ROBUST}                & 105.2 s \\
        192 / 192 & 0.01/255 & \texttt{ROBUST}                & 113.1 s \\
        192 / 192 & 5/255    & \texttt{NOT ROBUST} & 0 s     \\
        192 / 192 & 10/255   & \texttt{NOT ROBUST} & 0 s     \\
        192 / 192 & 1/255    & \texttt{Unknown}               & \bfseries\colorbox{pink}{7200 s}  \\
        192 / 192 & 0.05/255 & \texttt{ROBUST}                & \colorbox{green}{105.2 s}  \\
        \bottomrule
      \end{tabular}%
    }
\end{table}

\paragraph{Localized perturbations}
For sparse perturbations affecting a single pixel or ten pixels ($k \in \{1, 10\}$), both variants are certified \texttt{ROBUST} across all tested radii, with verification verdicts reached within approximately $100$--$130$ seconds. This indicates that the classifier is provably robust to the localized corruptions typically of small bubbles or sensor noise that affect only a few pixels of some region of interest.

\paragraph{Full-image perturbations}
When every pixel is perturbed ($k = 192$), the verdict depends strongly on the perturbation radius $\varepsilon$. At a small radius ($\varepsilon = 0.01/255$) both variants remain \texttt{ROBUST} (around $109$ and $113$ s), certifying stability against uniform low-intensity noise across the whole image. At large radii ($\varepsilon \in \{5/255, 10/255\}$) both variants are immediately reported \texttt{NOT ROBUST} as the falsification check finds a counter-example at essentially zero cost ($0$~s), and therefore no reachability computation is initiated. This highlights the practical value of the cheap falsification pre-check, which discards clearly non-robust cases without consuming much computation time.

\paragraph{The role of the activation functions}
At $\varepsilon = 1/255$ (full image attack), both variants return \texttt{UNKNOWN} after the allocated timeout of $7200$~s, meaning that the over-approximation is too loose to prove the robustness specification, and refinement does not close this gap within the fixed time limit. At the smaller radius $\varepsilon = 0.05/255$, the two activations diverge, as the Tanh variant is certified \texttt{ROBUST} in $105.2$~s, while the ReLU variant exhausts the allocated time duration. This contrast reflects a verifiability advantage of the smooth Tanh activation function dynamics, as its bounded derivative ($\tanh'(s) = \mathrm{sech}^2(s) \in (0,1]$) yields tighter Jacobian and mixed-monotonicity bounds and avoids the combinatorial case-splitting that the piecewise-linear ReLU induces in the star-set reachability of the post-ODE layers. 


\section{Conclusion and Future Work}
\label{Sect:conclusion}

We presented a reachability-based robustness verification framework for AUV-based plankton classifiers and introduced a compact neural ODE classifier for in-situ SilCam imagery. By combining mixed-monotonicity reachability for the continuous-time ODE block with star-set propagation for the surrounding layers, the framework provides formal guarantees that a classified image remains correctly classified under bounded $L_\infty$ perturbations, acting as an automated filter that reduces the manual post-processing burden on marine biologists. Our results further indicate that the Tanh activation function can be easier to verify than ReLU in the harder settings of full-image perturbation, which is a relevant observation when designing classifiers intended to be formally certified.

Future work includes extending the verification to higher input resolutions, deriving formal guarantees for the image downscaling stage preceding the classifier, and integrating the verifier into the onboard AUV processing pipeline for real-time filtering of ambiguous detections.

\section{Acknowledgements}
We acknowledge the support from Ahmed Abdelgayed throughout all phases of this work. We would like to thank Emlyn J. Davies and Raymond Nepstad from SINTEF Ocean for the initial discussions and feedback at the start of this work.


\bibliographystyle{IEEEtran}
\bibliography{REF}

\end{document}